\documentclass[a4paper]{article}

\usepackage{INTERSPEECH2021}

\usepackage{float}

\usepackage[all]{nowidow}

\usepackage{caption}
\usepackage{subcaption}

\usepackage{siunitx}

\usepackage{booktabs}

\usepackage{url}

\usepackage{color}
\definecolor{deepblue}{rgb}{0,0,0.5}
\definecolor{deepred}{rgb}{0.6,0,0}
\definecolor{deepgreen}{rgb}{0,0.5,0}

\usepackage{listings}
\usepackage{enumitem}
\setenumerate[1]{label={(\arabic*)}}

\usepackage[noadjust]{cite}

\usepackage{etoolbox}
\newcounter{BalanceAtReference}
\newcounter{ReferenceIndexForBalancing}
\makeatletter
\def\@balancelastpageonce{%
	\ifnum\value{ReferenceIndexForBalancing}=\value{BalanceAtReference}
	\newpage
	\else
	\relax
	\fi
	\stepcounter{ReferenceIndexForBalancing}
}
\pretocmd{\bibitem}{\@balancelastpageonce}
{} 
{\@latex@error{Patching \bibitem failed}{\@ehd}}
\makeatother

\title{Scribosermo: Fast Speech-to-Text models for German and other Languages}
\name{Daniel Bermuth, Alexander Poeppel, Wolfgang Reif}
\address{University of Augsburg, Institute for Software \& Systems Engineering}
\email{\{daniel.bermuth,alexander.poeppel,reif\}@informatik.uni-augsburg.de}

\begin{document}

\maketitle
  
\begin{abstract}
	Recent Speech-to-Text models often require a large amount of hardware resources and are mostly trained in English. This paper presents Speech-to-Text models for German, as well as for Spanish and French with special features:
	(a) They are small and run in real-time on microcontrollers like a RaspberryPi.
	(b) Using a pretrained English model, they can be trained on consumer-grade hardware with a relatively small dataset.
	(c) The models are competitive with other solutions and outperform them in German.
	In this respect, the models combine advantages of other approaches, which only include a subset of the presented features.
	Furthermore, the paper provides a new library for handling datasets, which is focused on easy extension with additional datasets and shows an optimized way for transfer-learning new languages using a pretrained model from another language with a similar alphabet.
\end{abstract}

\noindent\textbf{Index Terms}: fast speech to text, multilingual transfer-learning, automatic speech recognition, embedded hardware

\section{Introduction}
\label{sec:intro}

Speech-to-Text models based on neural networks are mostly trained in English and often require large amounts of training resources. But there exist many other languages and those who are interested in training a speech-to-text system for their own language do not always have access to high-performance server hardware. A few papers and projects focus on the aforementioned problems, but most are solving them only partially.

\vspace{9pt}
The authors of \textit{IMS-Speech} \cite{IMSDE} trained a German STT model, which so far had the best results on the German \textit{Tuda} dataset \cite{TUDA}. In a comparison with Google's STT service (executed 01/2019), their network could outperform it in English as well as in German.

In \textit{VoxPopuli} \cite{VOXPOP}, an approach for training multilingual models using a large unlabeled dataset is investigated. A mix of 50k hours of unlabeled data in different languages from the European Parliament and a comparatively small labeled dataset for semi-supervised training are used. This approach proved very effective and achieves a Word-Error-Rate (WER) of \SI{7.8}{\percent}\,$/$\,\SI{9.6}{\percent}\,$/$\,\SI{10.0}{\percent} in German\,$/$\,Spanish\,$/$\,French on the \textit{CommonVoice} datasets~\cite{COMV}, which so far have been the best results on these datasets.

\textit{Luo et al.} \cite{QNTECR} used the same network architecture as this work, but in Nvidia's original implementation, and also trained it for other languages like German or Spanish, using very small datasets and following a different transfer-learning approach of reinitializing the last network layer if the alphabet changes. 

Mozilla's \textit{DeepSpeech} project \cite{DEPSPE} provides pretrained English models that are relatively small and one of the few that are able to run in real-time on a RaspberryPi. It achieves a WER of \SI{7.1}{\percent} on the \textit{LibriSpeech} \cite{LIBSPE} testset. Some early experiments on multilingual trainings have been run with this network, but performance was much lower than the results presented in the following chapters. They still can be found in the project's repository which is linked later.

\textit{Park et al.} \cite{FNNMED} built a model for embedded devices, which reached a WER of \SI{9.0}{\percent} on \textit{LibriSpeech} and could run on an ARM-Cortex-A57. \textit{Zhang et al.} \cite{TINTRA} trained a very small model on a large in-house Chinese dataset which can run faster than real-time on an ARMv7 chip. \textit{He et al.} \cite{STRSRM} did train an English model on a very large in-house dataset which can run twice as fast than real-time on a Google-Pixel smartphone.

\textit{Ghoshal et al.} \cite{MLDNN} and \textit{Thomas et al.} \cite{MLFDNN} did run early explorations of transfer-learning for different languages using deep neural networks. The first approach replaces the last language specific layer of a network with a new one and finetunes the whole network on the new language, while the second uses a multilingual training of the first network layers, and different output layers for each language.

\vspace{9pt}
This paper presents a small Speech-to-Text model for German, as well as for Spanish and French, that combines the advantages of the aforementioned approaches. The main contributions of project \textit{Scribosermo} are:
(a) The models are competitive with the models from \textit{IMS-Speech} and \textit{VoxPopuli}.
(b) Providing pretrained models in multiple languages that can run in real-time even on single-board computers like a RaspberryPi.
(c) The models can be trained on a relatively small dataset, like the models from \textit{VoxPopuli} and only require consumer-grade hardware for training.
(d) Shows a fast transfer-learning approach with a single step through the concept of alphabet adaption.
(e) Improved SOTA performance for German STT.

Furthermore, the paper provides a new library for handling datasets, which is focused on easy extension with additional datasets and shows a simple way for transfer-learning new languages using a pretrained model from a language with an almost similar alphabet, which is demonstrated for English to Spanish and Spanish to Italian transfer-learning.

The training code and models are provided as open source at: \textit{https://gitlab.com/Jaco-Assistant/Scribosermo}

\noindent For reasons of readability, the results of the experiments are not presented in full detail, but can instead be found in the project's repository.

\section{Pre-processing}
\label{sec:prep}

The datasets are converted into single channel 16\,kHz audio with \textit{wav} encoding and a \textit{tab} separated \textit{csv} file for each partition, with at least the keys \textit{duration}, \textit{filepath} and \textit{text}. Afterwards an additional data cleaning step is executed.
All numbers are converted to their textual form, as well as commonly used units like \textit{kg} or \textit{m²}. After replacing some special characters (like \textit{ä}\textrightarrow{}\textit{ae}), all remaining characters, which do not match the used language's alphabet are removed. All of those rules are collected in a simple \textit{json} file, to ease adding new languages. 

In early training executions the transcriptions of some files did not match the recordings, which resulted in errors if they were much too short or much too long. Therefore, and in order to improve training speed, an automatic cleaning process was implemented, which excludes all files matching one of the following metrics:

\begin{enumerate}
	\itemsep0.1em
	\item Audio shorter than half a second.
	\item Audio longer than 30 seconds.
	\item Transcription has more than 512 characters.
	\item Recording is spoken 2x faster than the average.
	\item Less than one character per three seconds is spoken.
	\item (chars/second $<$ average$\times$3) and (duration $>$ average/5).
\end{enumerate}

The second and third items are used to exclude long files to allow for a greater training batch size. The forth and fifth metrics exclude too quickly or too slowly spoken utterances. The last is intended for slow recordings, too, but with an exception for short clips, because those may have longer pauses at the start or end of the recording.

\section{Language model}
\label{sec:langm}

To improve the predicted transcriptions of the trained network, the predictions are rescored with a 5-gram language model. For German a large 8-million sentence collection from \cite{SENDE} is used and combined with the transcriptions from the training dataset. The same text normalization steps as described in the last chapter are executed. In Spanish and French the \textit{Europarl} and \textit{News} sentence collections from \cite{STATMT} are used additionally, and the Italian training sentences are extended with the \textit{Mitads} dataset \cite{MITADS} The language model is created with \textit{Poco-LM} \cite{POCOLM} and optimized with tools provided by Mozilla's \textit{DeepSpeech} project \cite{DEPSPE}. For decoding their \textit{ds-ctcdecoder} is used as well. The language models were filtered to a maximum size of \mbox{165M n-grams}, which results in a size of about 850MB.

\section{Experiments with QuartzNet}
\label{sec:exqn}

\begin{figure}[htb]
	\centering
	\includegraphics[width=0.8\linewidth]{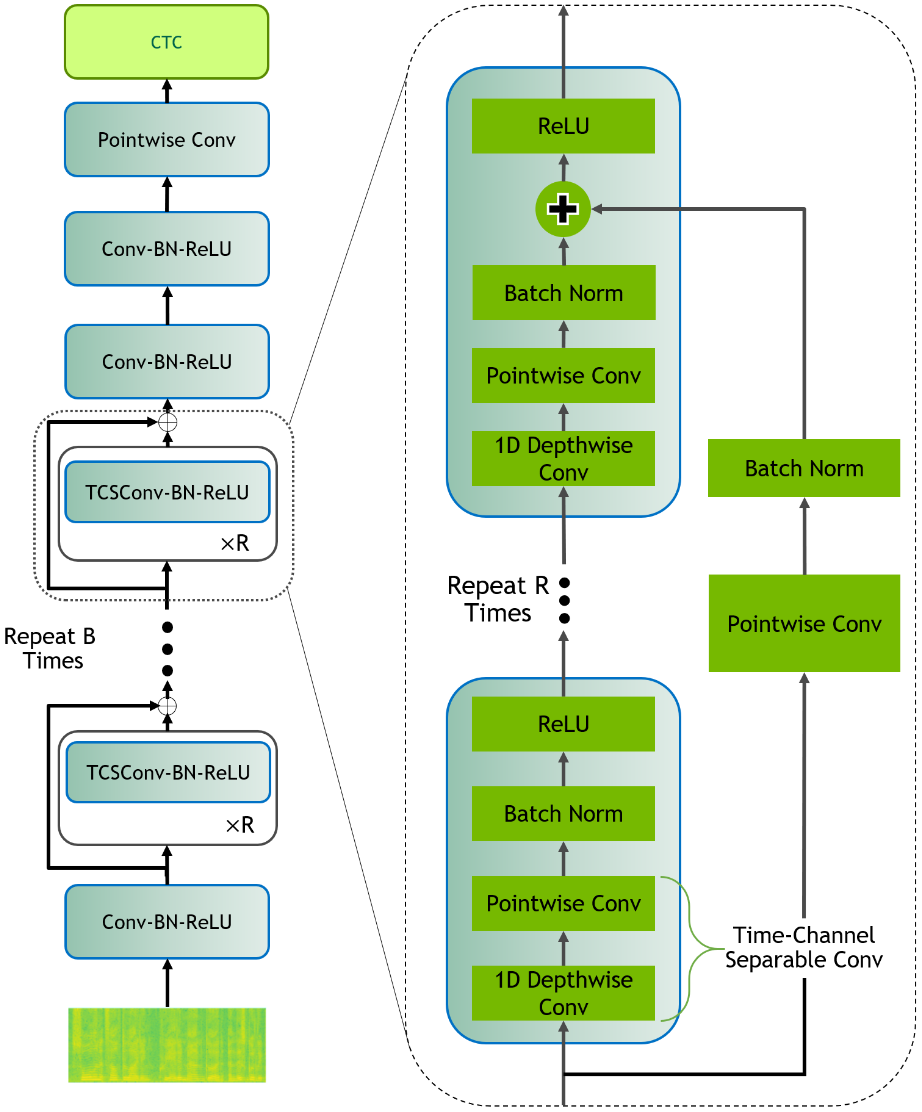}
	\caption{Network architecture of Nvidia's \textit{QuartzNet} \cite{QNET}.}
	\label{fig:qn}
\end{figure}

For the experiments the \textit{QuartzNet} architecture \cite{QNET} was implemented, using the open source code from Nvidia's \textit{NeMo} project \cite{NEMO} as reference.
The QuartzNet architecture (Figure~\ref{fig:qn}) was chosen, because its size is comparatively small, which results in fast inference on standard computers and low power devices. Nvidia provides pretrained weights for English, which reach a greedy WER of \SI{3.8}{\percent} on the LibriSpeech devset.

\subsection{Reimplementation in TensorFlow}

Instead of directly using Nvidia's PyTorch implementation, the network was reimplemented for TensorFlow. The main reason was that the trained network can be directly converted into the \textit{TensorFlow-Lite} format, which greatly improves inference speed on low power devices. Another benefit was that the tools already implemented for Mozilla's \textit{DeepSpeech} framework and some of their data augmentation features could be integrated more easily into the new project.

The pretrained weights from Nvidia's \textit{NeMo} project were transferred layer by layer from the PyTorch format to TensorFlow using \textit{Open Neural Network Exchange (ONNX)} as intermediate format.
While this did work well for the network itself, there were problems due to differences in PyTorch's and TensorFlow's spectrogram calculation. To reduce the impact of these, the transferred network was trained for four additional epochs on the LibriSpeech dataset. The performance is still slightly worse than Nvidia's reference implementation, but much better than Mozilla's current \textit{DeepSpeech} release (Table~\ref{tab:qen}).

\begin{table}[H]
	\caption{Performance on English LibriSpeech dataset. Sometimes predictions are rescored with an additional 5-gram language model (LM), else the greedy WER is measured.}
	\label{tab:qen}
	\centering
	\begin{tabular}{llc}
		\toprule
		\textbf{Network} & \textbf{Notes} & \textbf{WER} \\
		\midrule
		DeepSpeech & LS-test-clean + LM & \SI{7.06}{\percent} \\
		\midrule
		QuartzNet15x5 & Nvidia, LS-dev-clean & \SI{3.79}{\percent} \\
		QuartzNet15x5 & Converted, LS-dev-clean & \SI{5.15}{\percent} \\
		QuartzNet15x5 & Trained, LS-dev-clean & \SI{4.35}{\percent} \\
		QuartzNet15x5 & above, LS-test-clean & \SI{4.57}{\percent} \\
		QuartzNet15x5 & above, LS-test-clean + LM & \SI{3.71}{\percent} \\
		\bottomrule
	\end{tabular}
\end{table}

\subsection{Training in German}

For the following trainings the \textit{CommonVoice} (v6) dataset was used. The full training partitions are larger than the amount used in \textit{VoxPopuli} \cite{VOXPOP}, therefore a random subset was selected to match the overall duration.
In order to get the same alphabet as in English, the German umlauts (\textit{ä},\,\textit{ö},\,\textit{ü}) have been replaced with their transliteration (\textit{ae},\,\textit{oe},\,\textit{ue}).

Table~\ref{tab:rqn} shows that the implemented \textit{QuartzNet15x5} network and training procedure, named \textit{Scribosermo} in the table, can outperform other approaches for German speech recognition. The training setup of simply training over a pretrained English network, without any further changes, is straightforward, and does not require a semi-supervised pretraining on multiple languages.

\begin{table}[!htbp]
	\caption{German training results. Above networks have been tested on CommonVoice, below on Tuda dataset.}
	\label{tab:rqn}
	\centering
	\begin{tabular}{llcc}
		\toprule
		\textbf{}  & \textbf{Notes}  & \textbf{Duration} & \textbf{WER} \\
		\midrule
		Scribosermo & using CV v6 & 314h  & \SI{7.7}{\percent} \\
		VoxPopuli \cite{VOXPOP} & using CV v5 & 314h &
		\SI{7.8}{\percent} \\
		Luo et al. \cite{QNTECR} & greedy on devset & 119h &
		\SI{18.7}{\percent} \\
		\midrule
		Scribosermo & above model & 314h & \SI{11.7}{\percent} \\
		IMS-Speech \cite{IMSDE} & mixed dataset & 806h & \SI{12.0}{\percent} \\
		\bottomrule
	\end{tabular}
\end{table}

In Table~\ref{tab:tdts} some different training modalities are investigated. The first section uses the complete German training partition of \textit{CommonVoice}, which slightly improves the results. This training took 3~days on a PC with two \mbox{1080Ti GPUs}, which shows that the training process itself is very fast, too, and can be executed on consumer-grade hardware.
The second part shows that training results can be improved by a larger margin if the training is run again with the same parameters, but using the current model checkpoint as initialization model. This follows the ideas of \textit{Stochastic Gradient Descend with Restart} \cite{SGDR}, but uses the already implemented early-stopping with learning rate reductions on plateaus approach instead of a cosine annealing learning rate.

\begin{table}[H]
	\caption{Testing different training setups.}
	\label{tab:tdts}
	\centering
	\begin{tabular}{lcc}
		\toprule
		\textbf{Notes}  & \textbf{Duration} & \textbf{WER} \\
		\midrule
		full CV trainset & 720h  & \SI{7.5}{\percent} \\
		\midrule
		Iteration 1 & 314h & \SI{8.3}{\percent} \\
		Iteration 2 & 314h & \SI{7.8}{\percent} \\
		Iteration 3 & 314h & \SI{7.7}{\percent} \\
		\bottomrule
	\end{tabular}
\end{table}

\subsection{Training in other languages}

\textit{Scribosermo's} approach is competitive in other languages like Spanish and French as well, which is shown in Table~\ref{tab:rqnef}. To simplify the transfer-learning process the usual two-step frozen and unfrozen training with a reinitialized last layer was replaced with a simpler approach that does not require freezing of parts of the network.
First, the alphabet size of the two languages was reduced, using the rules for cross-word puzzles, which replace letters that contain diacritics and ligatures with their basic form. Using the cross-word puzzles rules has the advantage that they are commonly known and therefore should not pose a problem for humans reading the predicted transcriptions. Following this approach, the French alphabet now has the same letters as the English, only the Spanish has an extra letter\;(ñ). Thus, the size of the last layer for Spanish still has to be changed, but instead of completely reinitializing the new layer, it is only extended with new weights for the extra letter. Thereby the pretrained English weights for the other letters can be kept, which greatly improves the results, similar to only training over the pretrained weights in German. A future optimization step could include replacing phonetically similar but otherwise different characters in the base alphabet with ones from the target alphabet, which was explored in more depth by \cite{ZRTCSW} for training-free language adoption.

\begin{table}[!htbp]
	\caption{Spanish and French training results on CommonVoice testset. Above is Spanish, below is French.}
	\label{tab:rqnef}
	\centering
	\begin{tabular}{llcc}
		\toprule
		\textbf{}  & \textbf{Notes}  & \textbf{Duration} & \textbf{WER} \\
		\midrule
		Scribosermo & using CV v6 & 203h  & \SI{10.9}{\percent} \\
		VoxPopuli \cite{VOXPOP} & using CV v5 & 203h & \SI{9.6}{\percent} \\
		Luo et al. \cite{QNTECR} & greedy on devset & 96h &
		\SI{15.0}{\percent} \\
		\midrule
		Scribosermo & using CV v6 & 364h & \SI{12.5}{\percent} \\
		VoxPopuli \cite{VOXPOP} & using CV v5 & 364h & \SI{10.0}{\percent} \\
		\bottomrule
	\end{tabular}
\end{table}

\vspace{9pt}
To compare the influence of the alphabet extension, a separate experiment was run following the usual training approach of reinitializing the complete last layer for the larger Spanish alphabet. The first part of Table~\ref{tab:iae} shows that the single-step approach with alphabet extension performs better than a simple single-step training with reinitialization of the last layer, as the deeper layers' weights are not so much influenced by backpropagation of prediction errors coming from the random weights of the last layer at the beginning of the training. Compared to the more usual two-step training which solves this problem it is much faster (trainings were executed on $2\times$ Nvidia-V100). 

Similar to extending the last layer of the network for new alphabet letters, it is also possible to drop characters to reduce the alphabet size. Following the cross-word puzzle approach, the converted Italian alphabet is the same as the English one. But as Italian sounds more similar to Spanish than to English, it is beneficial to use a Spanish network to train upon, after dropping the extra Spanish letter (Table~\ref{tab:iae}, second part).

\begin{table}[H]
	\caption{Influence of finetuning with alphabet extension on Spanish (above) and alphabet shrinking for Italian (below).}
	\label{tab:iae}
	\centering
	\begin{tabular}{lcc}
		\toprule
		\textbf{Notes} & \textbf{WER} & \textbf{Traintime} \\
		\midrule
		single-step reinitialization & \SI{11.66}{\percent} & 18h \\
		two-step training & \SI{11.11}{\percent} & 13+18h \\
		
		alphabet extension & \SI{11.05}{\percent} & 19h \\ 
		\midrule
		\textbf{Notes}  & \textbf{Duration} & \textbf{WER} \\
		\midrule
		English $\rightarrow$ Italian & 111h  & \SI{13.8}{\percent} \\
		Spanish $\rightarrow$ Italian & 111h  & \SI{12.2}{\percent} \\
		\bottomrule
	\end{tabular}
\end{table}

\subsection{Inference speed}
The main benefit of a small network is fast inference speed on low powered devices. This usually comes with a trade-off of a loss in recognition accuracy as larger models can store more information in their weights, but a fast model that can run even on devices with low computation capabilities is a key feature of this work. The full model itself has a size of about 75MB, the quantized model about 20MB.

The transcription speed is evaluated in Table~\ref{tab:insp}, which shows that the presented models are much faster than the model of \textit{IMS-Speech}, and that they can run faster than real-time on a RaspberryPi. To reduce the memory requirements for very long inputs, they can also be transcribed chunk by chunk in a streaming manner. Here the full CTC-labels were calculated and afterwards given as input to the decoder, similar as in \textit{IMS-Speech}.
The authors of \textit{VoxPopuli} did not publish the inference speed of their network. A comparison of the network parameter count between their \textit{wav2vec-base} net, which has about 95M params, and \textit{QuartzNet15x5} which only has 19M, allows an estimation that it might run about 5x slower.

\begin{table}[H]
	\caption{Inference Speed, measured as Real Time Factor}
	\label{tab:insp}
	\centering
	\begin{tabular}{llc}
		\toprule
		\textbf{Device} & \textbf{Model} & \textbf{RTF} \\
		\midrule
		PC - 1 core AMD3700X &  & $0.24$ \\
		PC - 1 core (unknown) & net of \textit{IMS-Speech} \cite{IMSDE} & $14.2$ \\
		\midrule
		RaspberryPi-4 - 4gb & tflite full & $1.3$ \\
		RaspberryPi-4 - 4gb & tflite optimized (TLO) & $0.7$ \\
		RaspberryPi-4 - 4gb & \textit{DeepSpeech} TLO & $0.7$ \\
		\bottomrule
	\end{tabular}
\end{table}

\subsection{Training with all datasets}

After demonstrating the performance of the presented approach with relatively small datasets, an experiment was run to measure the influence of larger datasets on the transcription performance.
In total 37~datasets for German \cite{VOXF,TUDA,COMV,FRMT,SPRIN,CSS10,GOTH,KURZG,LINGL,MUWI,MALBS,PULSR,SWC,TATO,TERAX,YKOLL,ZAMSP,ALC,BROTH,HMPL,PHATT,PD1,RVG1,RVGJ,SC10,SHC,SI100,SMC,VM1,VM2,WASEP,ZIPTEL,TORST,GLD2,SKYRM,WIT3,MTEDX}, 8~datasets for Spanish \cite{VOXF,COMV,CSS10,LINGL,MALBS,TATO,MTEDX,LVES}, 7~datasets for French \cite{VOXF,COMV,CSS10,LINGL,MALBS,TATO,MTEDX} and 5~datasets for Italian \cite{VOXF,COMV,LINGL,MALBS,MTEDX} were collected. The trainings were continued with the models of the trainings with CommonVoice only, and afterwards the models were finetuned on this dataset again. The results can be found in Table~\ref{tab:tads} and show that the advantage of using more data is relatively small. Possible explanations might be that the quality of the mixed datasets is not very good or differs too much from the test recordings, or that the small network is reaching its maximum information capacity.

\begin{table}[!htbp]
	\caption{Training with all accessible datasets  in German (DE), Spanish (ES), French (FR) and Italian (IT). Datasets for testing are either CommonVoice (CV) or Tuda (TD).}
	\label{tab:tads}
	\centering
	\begin{tabular}{llcc}
		\toprule
		\textbf{Language} & \textbf{\#Datasets} & \textbf{Duration} & \textbf{WER} \\
		\midrule
		DE-CV & $37$ & \SI{2370}{\hour}  & \SI{6.6}{\percent} \\
		DE-TD &  &   & \SI{10.2}{\percent} \\
		\midrule
		ES-CV & $8$ & \SI{817}{\hour}  & \SI{10.0}{\percent} \\
		FR-CV & $7$ & \SI{1028}{\hour}  & \SI{11.0}{\percent} \\
		IT-CV & $5$ & \SI{360}{\hour}  & \SI{11.5}{\percent} \\
		\bottomrule
	\end{tabular}
\end{table}

\section{Corcua}
\label{sec:corcua}

In this chapter the library which was built to handle the above datasets is presented.
Often speech-to-text frameworks like Mozilla's \textit{DeepSpeech} or Nvidia's \textit{NeMo} have customized scripts for downloading and converting a range of supported datasets into their custom dataset format. But many datasets are used in multiple frameworks, so large parts of the scripts have overlapping tasks.
The \textit{audiomate} \cite{AUDMAT} library was built to ease the use of different audio datasets for machine learning tasks and is able to load 18 different datasets and export them into 4 different speech-to-text frameworks. But extending it with new datasets is quite complicated and requires a deeper understanding of the architecture.
The goal in creating \textit{corcua} was not only to build a library to load different audio datasets and export them to different framework formats, but also to make adding new datasets as easy as possible.

\vspace{9pt}
\textit{Corcua's} architecture is split into three different parts, \textit{downloading}, \textit{reading} and \textit{writing}.
The \textit{downloader's} task is to download and extract a dataset to a local directory. Helper functions for common formats like \textit{zip} and \textit{tar.gz} or downloading from a server directory are already pre-implemented.
A \textit{reader} loads the audio files, transcriptions and optionally other information included in the dataset into an easy to handle dictionary format and returns a list of items like this:

\noindent
\begin{minipage}{\linewidth}
\begin{lstlisting}
item = {
  "filepath": "path/to/audiofile",
  "speaker": "Newton",
  "text": "That damn apple!"
}	
\end{lstlisting}
\end{minipage}

A \textit{writer} takes a list of dictionaries and saves them into the requested framework's dataset format, like \textit{csv} or \textit{json}. It also converts the audio files from different codecs to the commonly used \textit{wav} encoding.

Besides mere dataset processing, there are also tools to print some statistics about the dataset, like total duration or the most recorded speakers. \textit{Corcua} also supports splitting datasets into different partitions, like train and test, either randomly or by key separated classes, for example that all utterances of one speaker are in the same partition.

\vspace{9pt}
Compared to \textit{audiomate}, conversions of some datasets are much faster. Converting the German CommonVoice-v5 dataset~(\SI{701}{\hour}) with \textit{audiomate} took about \SI{12}{\hour} on a modern CPU, while converting the slightly larger CommonVoice-v6 dataset~(\SI{777}{\hour}) with \textit{corcua} takes less than \SI{3}{\hour}. 

Currently \textit{corcua} can load 34 different datasets (18 of them are German only, 13 are available in more than three languages), and is able to write them into 3 framework formats.
Some of the multilingual datasets have been extracted from computer games like \textit{Skyrim} or \textit{The\,Witcher}, which often provide high quality dialogs. With the included support of extracting labels from  manually transcribed YouTube videos, it is possible to create audio datasets for almost any language.
\textit{Corcua} has been released as open source project and can be accessed under: \textit{https://gitlab.com/Jaco-Assistant/corcua}

\section{Conclusion}
\label{sec:conclu}

In this paper small Speech-to-Text models for German, as well as for Spanish, French and Italian, were presented. The models combine the advantages of other approaches. They are competitive with the best models to date on the CommonVoice dataset in German, Spanish and French, as well as with the best one on the German Tuda dataset. At the same time they can run in real-time on single-board computers like a RaspberryPi and can be trained on consumer-grade hardware with a comparatively small dataset.
These models are especially interesting for embedded or offline speech applications, for example in smart home systems running on edge-devices with low power consumption, or on smartphones in environments where no stable internet connection is available. Running offline on standard hardware also has advantages if users do not want and companies are not allowed to use cloud providers for privacy reasons. 

\bibliographystyle{IEEEtran}
\bibliography{mybib}

\end{document}